\documentclass[runningheads]{llncs}
\usepackage[T1]{fontenc}
\usepackage{graphicx,verbatim}
\usepackage{amsmath,amsfonts}
\usepackage{algorithmic}
\usepackage{algorithm}
\usepackage{array}
\usepackage[caption=false,font=normalsize,labelfont=sf,textfont=sf]{subfig}
\usepackage{textcomp}
\usepackage{stfloats}
\usepackage{url}
\usepackage{verbatim}
\usepackage{graphicx}
\usepackage{cite}
\usepackage{bbm}
\usepackage{bm}
\usepackage{csquotes}
\usepackage{multirow}
\usepackage{marvosym}
\usepackage{booktabs}
\usepackage[colorlinks=true, urlcolor=blue]{hyperref}
\usepackage{amssymb}

\begin{document}
\title{CLIMD: A Curriculum Learning Framework for Imbalanced Multimodal Diagnosis}
  
\author{Kai Han 
\and Chongwen Lyu 
\and Lele Ma 
\and Chengxuan Qian 
\and
\\Siqi Ma 
\and Zheng Pang 
\and Jun Chen 
\and Zhe Liu \textsuperscript{\Letter} 
}

\authorrunning{Kai Han et al.}
%
\institute{School of Computer Science and Telecommunication Engineering, \\Jiangsu University, China \\
\email{1000004088@ujs.edu.cn}}

\maketitle         
\begin{abstract}
Clinicians usually combine information from multiple sources to achieve the most accurate diagnosis, and this has sparked increasing interest in leveraging multimodal deep learning for diagnosis. However, in real clinical scenarios, due to differences in incidence rates, multimodal medical data commonly face the issue of class imbalance, which makes it difficult to adequately learn the features of minority classes. Most existing methods tackle this issue with resampling or loss reweighting, but they are prone to overfitting or underfitting and fail to capture cross-modal interactions. Therefore, we propose a \textbf{C}urriculum \textbf{L}earning (CL) framework for \textbf{I}mbalanced \textbf{M}ultimodal \textbf{D}iagnosis (CLIMD). Specifically, we first design multimodal curriculum measurer that combines two indicators, intra-modal confidence and inter-modal complementarity, to enable the model to focus on key samples and gradually adapt to complex category distributions. Additionally, a class distribution-guided training scheduler is introduced, which enables the model to progressively adapt to the imbalanced class distribution during training. Extensive experiments on multiple multimodal medical datasets demonstrate that the proposed method outperforms state-of-the-art approaches across various metrics and excels in handling imbalanced multimodal medical data. Furthermore, as a plug-and-play CL framework, CLIMD can be easily integrated into other models, offering a promising path for improving multimodal disease diagnosis accuracy. Code is publicly available at https://github.com/KHan-UJS/CLIMD. 


\end{abstract}
\section{Introduction}

Thanks to the rapid development of deep learning, Computer-Aided Diagnosis (CAD) has experienced a new era of advancement. Traditional deep diagnosis researches mostly focus on a single modality \cite{liu2022improving, textreview3, tablereview1, hu2024imbalance,han2025region}, however, diagnosing diseases based on only one modality is challenging and risky. Therefore, in recent years, the application of multimodal deep learning (MMDL) in disease diagnosis, which makes comprehensive judgments by combining information from multiple modalities, has garnered widespread attention \cite{MMDLreview1, MMDLreview3, MMDLreview4,xing2025re,qian2025decalign}.

In real medical cases, common diseases usually constitute a significant proportion, while samples of some rare malignant diseases are relatively scarce, leading to the issue of class imbalance. This issue makes it challenging for the model to learn the minority classes, which can lead to biased predictions. There is relatively limited research addressing the class imbalance issue in multimodal diagnosis (MD). Some existing methods attempt to apply common resampling \cite{resample1}, which carries the risk of overfitting or underfitting, as well as loss reweighting \cite{reweight1, reweight2, reweight3}, which may increase the instability of model training. Additionally, these methods overlook the impact of interactions between modalities in MMDL.

To this end, we propose a MD framework based on curriculum learning (CL) \cite{CL1,qian2025dyncim} named CLIMD. CL enables the model to transition gradually from simple to complex during the training process \cite{CL2}.
Inspired by the concept of CL, we attempted to quantify the degree of class imbalance in the dataset and then train the model using a progressive sampling strategy. Specifically, we first propose a curriculum measurer combining intra-modal confidence and inter-modal complementarity, to evaluate the training difficulty of each multimodal sample. Based on the difficulty, a class distribution-guided curriculum scheduling algorithm is designed to quantify the degree of class imbalance and construct training subsets with gradually increasing class imbalance in each epoch, helping the model adapt to imbalanced data. CLMD does not involve any down-sampling, data generation, or data augmentation operations, thereby avoiding the issues of underfitting or overfitting arise from them. The main contributions of this paper can be summarized as follows:
\begin{itemize}
    \item We propose a novel framework based on curriculum learning to mitigate the issue of class imbalance in multimodal diagnosis. To the best of our knowledge, this is the first CL strategy specifically designed for imbalanced multimodal diagnosis.
    
    \item We design a curriculum measurer and a class distribution-guided training scheduler to measure the training difficulty of multimodal samples and help the model gradually adapt to the imbalance during training.
    
    \item Extensive experiments conducted on multiple datasets demonstrate that the proposed method significantly outperforms state-of-the-art methods and can effectively address the issue of class imbalance.

\end{itemize}

\section{Method}

\begin{figure*}[!t]
	\centering
	\includegraphics[width=0.85 \linewidth]{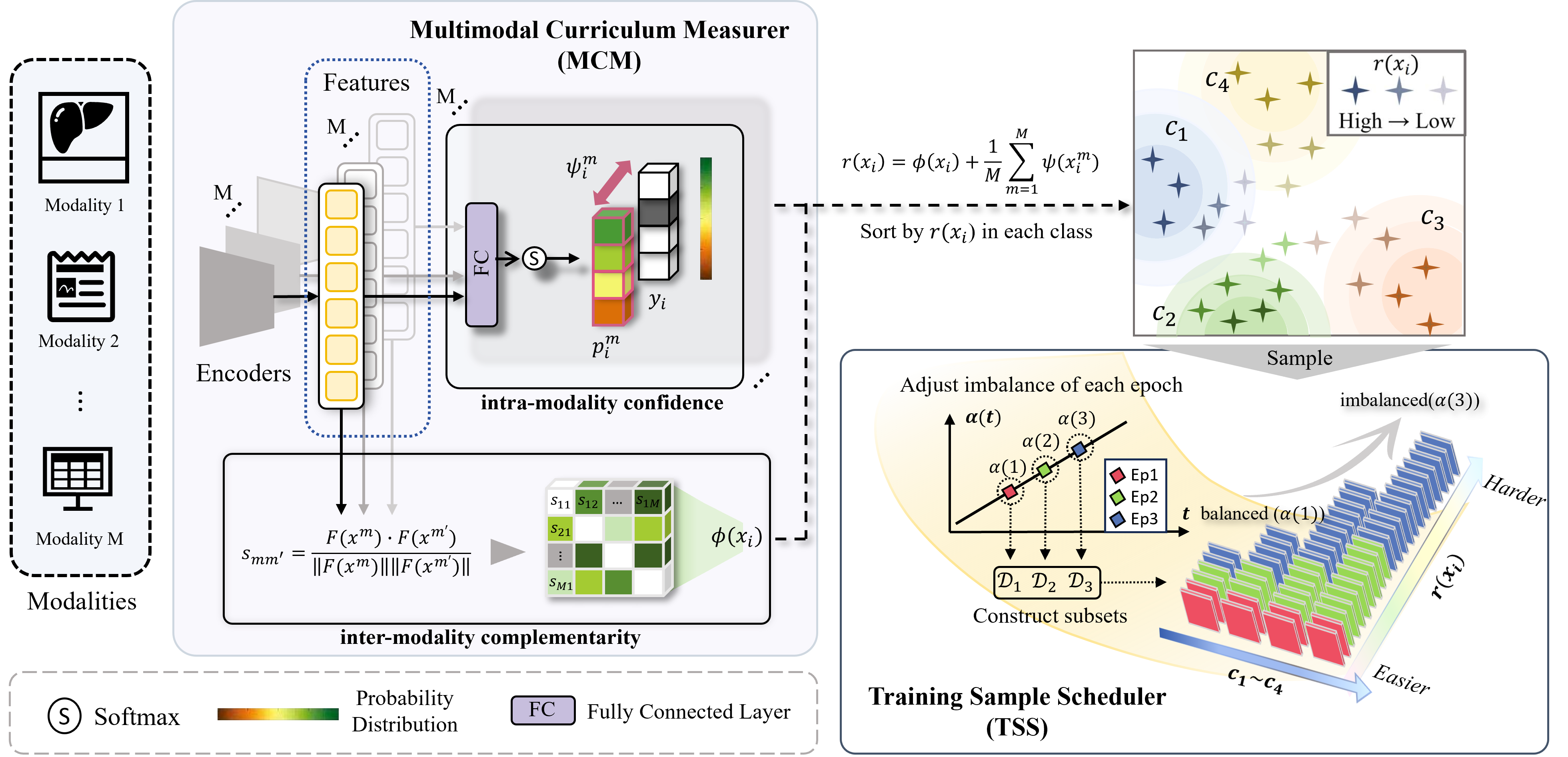}
	\caption{Overview of CLIMD, which includes Multimodal Curriculum Measurer (MCM) and Training Sample Scheduler (TSS). (i) MCM evaluates the training difficulty of each sample from intra-modal confidence $\psi_i^{(m)}$ and inter-modal complementarity $\phi(x_i)^{(m)}$, which are combined to calculate the overall training difficulty $r(x_i)$ for each sample. (ii) TSS samples from each class in order to construct the training subset $\mathcal{D}_t$ for the $t$-th epoch and gradually increases the class imbalance degree in the training subset by adjusting $\alpha(t)$.}
	\label{fig_1}
\end{figure*}

In this section, we introduce the proposed method in detail, including 
a Multimodal Curriculum Measurer (MCM) and a Training Sample Scheduler (TSS). The overview of CLIMD is shown in Fig. \ref{fig_1}. It is noteworthy that, to evaluate the upper bound of the proposed CL without interference from other factors, the multimodal diagnosis (MD) network we employ (i.e., \textbf{Baseline}) is a basic and classical early fusion architecture with Cross-Entropy (CE) as the loss function. In fact, CLIMD is plug-and-play and can be applied to other more complex MD models.

\subsection{Definition}
Given a dataset $\mathcal{D}$ with $M$ modalities, which includes $N$ multimodal data observations with corresponding labels, i.e., $\mathcal{D}=\{\{x_i^{(m)}\}_{m=1}^M,y_i\}_{i=1}^{N}$. The main goal of the MD task is to learn a mapping from the multimodal data space $\bm{X}=\{x_i^{(m)}\in \mathbb{R}^{d_m}\}_{m=1}^M$ to the label space $\bm{Y}=\{y_i\in \mathbb{R}^{C}\}$, where $d_m$ and $C$ are the dimensionality of $M$ modalities and class number, respectively.

\subsection{Multimodal Curriculum Measurer} \label{MCM}
In imbalanced multimodal learning, random sampling may lead to model bias toward majority classes or insufficient consideration of modality interactions. To sample in a specific order, we propose the multimodal curriculum measurer as a structured evaluation criterion to regulate the training process. The Multimodal Curriculum Measurer (MCM) comprises intra-modal confidence, which prioritizes stable samples to enhance convergence robustness, and inter-modal complementarity, which facilitates the learning of complementary information across modalities to improve fusion effectiveness.

\subsubsection{Intra-modal Confidence}
The prediction confidence of a sample in its $m$-th modality quantifies the model's certainty regarding the true class. To compute this, the class probability output $p_{i,c}^{m}$ and the true class label $y_{i,c}$ are utilized to derive the confidence score $\psi(x_i^{m})$, as defined in Eq. \ref{eq1}. 
\begin{equation}
	\label{eq1}
	\psi(x_i^{m}) = \sigma\left[\frac{1}{C}\sum_{c=1}^{C}y_{i,c} \log p_{i,c}^{m}\right]
\end{equation}
where $\sigma$ denotes the sigmoid function.

\subsubsection{Inter-modal Complementarity}
Inter-modal complementarity quantifies the complementarity score by computing modality feature similarity. Specifically, we extract modality-specific representations $F(x^m)$ through encoders, then compute pairwise modality feature similarity $s_{mm^{'}}$ to obtain the similarity matrix 
$\mathcal{S}$. $s_{mm^{'}} \in \mathcal{S}$, where $m\in \{1,\cdots,M\}$, $m^{'}\in \{1,\cdots,M\}$. The calculation process of feature similarity $s_{mm^{'}}$ is as follows:
\begin{equation}
\label{eq4}
s_{mm'} = \frac{F(x^m) \cdot F(x^{m'})}{\Vert F(x^m)\Vert \Vert F(x^{m'})\Vert}
\end{equation}
Based on the similarity matrix $\mathcal{S}$, modality complementarity is computed as shown in the following equation.
\begin{equation}
	\label{eq5}
	\phi(x_i) = 1-\frac{1}{M(M-1)}\sum_{m\neq m^{'}}s_{mm'}
\end{equation}

Finally, the intra-modal confidence and inter-modal complementarity scores are integrated, as shown in Eq. \ref{eq6}.
\begin{equation}
	\label{eq6}
	r(x_i) = \phi(x_i)+\frac{1}{M}\sum_{m=1}^{M}\psi(x_i^{m})
\end{equation}

\subsection{Class Distribution-guided Training Scheduler}
To help the model gradually adapt to imbalance, we propose a Training Sample Scheduler (TSS) guided by class distribution. TSS first quantifies the degree of class imbalance in the dataset. Then, the model starts learning from a subset with relatively balanced distribution, and progressively increases the degree of class imbalance until all imbalanced data is involved in the training process.

\subsubsection{Fit the Class Distribution}
We assume that there are $C$ classes in the dataset $\mathcal{D}$, i.e., $\mathcal{D}=\{\mathcal{D}_{sub}^{(c)}\}_{c=1}^C$. A discrete class distribution $N=\{n^{(c)}\}_{c=1}^C$ can be obtained, where $n^{(c)}$ represents the number of samples in the $c$-th class. Next, we fit $N$ to an improved power-law distribution, whose general probability density function (PDF) is as follows:
\begin{equation}
\label{eq8}
    P(N)=(\gamma\alpha-1){n_{min}^{\gamma\alpha-1}}N^{-\gamma\alpha}
\end{equation}
where $n_{min}$ represents the smallest value in $N$, $\gamma>0$ is a hyperparameter that makes the distribution function smoother, and $\alpha>1/\gamma$ is the parameter to be estimated. The optimal solution for $\alpha$ is obtained through Maximum Likelihood Estimation (MLE) as:
\begin{equation}
    \hat{\alpha}={\frac{1}{\gamma}}\left(1+\frac{C}{\sum_{c=1}^C\ln{n^{(c)}}-C\ln{n_{min}}}\right)
\end{equation}
Notably, $\hat{\alpha}$ here can be regarded as the upper bound of the class imbalance degree, and it is written as $\alpha(T)$ in the following content.

\subsubsection{Construct the Training Subsets}

\begin{figure}[!t]
	\centering
	\includegraphics[width=.6 \linewidth]{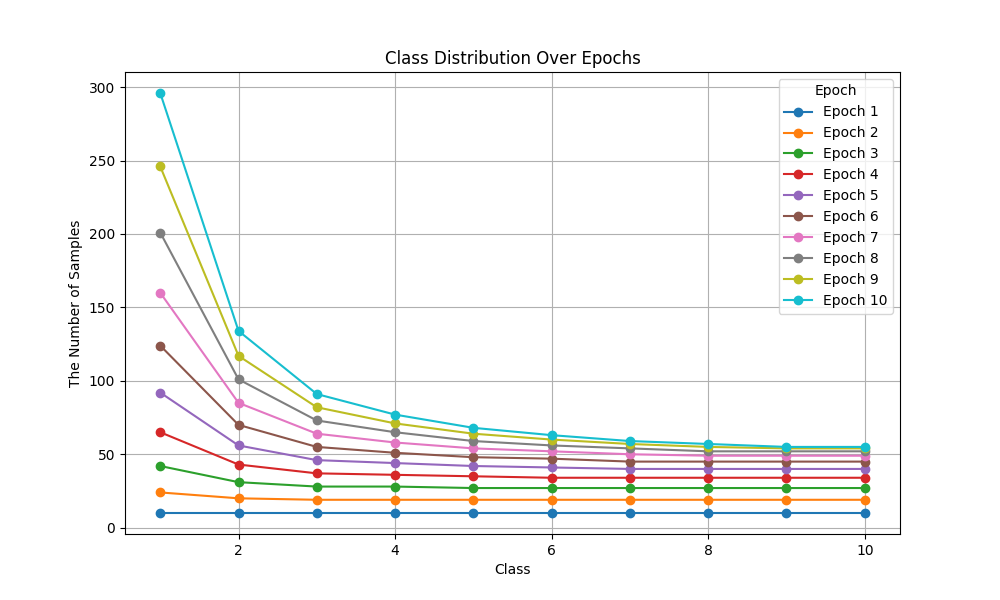}
	\caption{An example of class distribution over epochs, where the sample number $N=1000$, the total epochs $T=10$, the number of classes $C=10$, the hyperparameter of the real class distribution (in Epoch 10) function $\alpha(T)=5$.}
	\label{fig_2}
\end{figure}

According to training difficulty $r(x_i)$ obtained in Section \ref{MCM}, all samples within their respective classes are sorted from easy to hard, waiting to be sampled in order later. We assume that the total number of training epochs is $T$, and the training subset for the $t$-th epoch is $\mathcal{D}_t$, where $t\in[0,T]$. The number of samples in each training subset is $S_t=\left\{s_t^{(c)}\right\}_{c=1}^C=\frac{t}{T}\cdot N$, where $s_t^{(c)}$ represents the number of samples in the $c$-th class. In the first training subset $\mathcal{D}_1$, the classes follow a uniform distribution, meaning that the number of samples for each class is $S_1/C$. Next, we control the degree of class imbalance in each epoch by adjusting $\alpha(t)=1+\left(\alpha(T)-1\right)\cdot\frac{t-1}{T-1}$, and $1\leq \alpha(t) \leq \alpha(T)$. Therefore, the class distribution function of the training subset in the $t$-th epoch is a mixture of uniform distribution and power-law distribution, as follows:
\begin{equation}
    q_t(c)=\left(1-\frac{t-1}{T-1}\right) \cdot \frac{1}{C}+ \left(\frac{t-1}{T-1}\right) \cdot \frac{c^{-\gamma\alpha(t)}}{\sum_{i=1}^{C}i^{-\gamma\alpha(t)}}
\end{equation}
where $q_t(c)$ represents the sampling probability of $c$-th class in the $t$-th epoch, and the number of samples drawn from the $c$-th class in order is $s_t^{(c)}=q_t(c) \cdot S_t$.

To provide an intuitive illustration, we have included an example in Fig. \ref{fig_2}, where $N=1000$, $T=10$, $C=10$, $\alpha(T)=5$. It shows that the proposed scheduling strategy allows the training subsets to gradually transition from a uniform distribution to a long-tail distribution (which approximates real imbalanced class distribution), enabling the model to progressively adapt to the class imbalance in the training samples.

\subsection{Experiment}
\subsubsection{Datasets and Evaluation Metrics} Multimodal Liver Lesion Classification (MLLC) is a private multimodal liver lesion diagnosis dataset containing 320 patients, involving three classes: normal, metastatic liver lesions, and hepatocellular carcinoma. It consists of CT arterial phase images, medical history texts, and laboratory test indicators.  Breast Invasive Carcinoma (BRCA) is a publicly available dataset in The Cancer Genome Atlas (TGCA) \footnote{\href{https://www.cancer.gov/ccg/research/genome-sequencing/tcga}{https://www.cancer.gov/ccg/research/genome-sequencing/tcga}} program. It is used for breast invasive carcinoma subtype classification, containing 3 modalities: mRNA, DNA methylation, and miRNA.

Three metrics are used to quantify the experimental results: accuracy (ACC), weighted F1 score, and macro F1 score.

\subsubsection{Implementation Details}
For the MLLC dataset, a pretrained ResNet18 \cite{resnet} is used to extract features from CT. In addition, we use MedBERT \cite{medbert} to encode medical texts and a two-layer fully connected network (FCN) to encode the tabular data. For the BRCA dataset, we preprocess it referring to \cite{mogonet} \footnote{Code is available at \href{https://github.com/txWang/MOGONET}{https://github.com/txWang/MOGONET}}. The learning rate is set to 5e-5, the training epochs $T$ is 100, the hyperparameter $\gamma$ is empirically set to 0.3. All training processes are performed on an NVIDIA GeForce RTX A6000 GPU.

\begin{table*}[htbp]
\caption{The comparisons with SOTA MD methods on MLLC and BRCA.}
\label{tab1}
    \begin{center}
    \resizebox{12cm}{!}{
    \begin{tabular}{l|ccc|ccc}
        \toprule
        \multirow{2}{*}{\textbf{Method}} & \multicolumn{3}{c|}{\textbf{MLLC}} & \multicolumn{3}{c}{\textbf{BRCA}} \\
        \cline{2-7}
        & ACC & Weighted F1 & Macro F1 & ACC & Weighted F1 & Macro F1 \\
        \hline
        MORONET \cite{mogonet} (NC' 2021) & 72.2 $\pm$ 4.6 & 67.2 $\pm$ 4.9 & 68.3 $\pm$ 6.8 & 82.9 $\pm$ 1.8 & 82.5 $\pm$ 1.6 & 77.4 $\pm$ 1.7 \\
        Multi-D \cite{dynamicalfusion} (CVPR' 2022) & 74.4 $\pm$ 5.0 & 68.1 $\pm$ 4.6 & 70.3 $\pm$ 4.2 & 87.7 $\pm$ 0.3 & 88.0 $\pm$ 0.5 & 84.5 $\pm$ 0.5 \\
        MLCLNet \cite{MLCLNet} (AAAI' 2023) & 76.4 $\pm$ 3.5 & 74.5 $\pm$ 3.2 & 72.6 $\pm$ 3.7 & 86.4 $\pm$ 1.6 & 87.8 $\pm$ 1.7 & 82.6 $\pm$ 1.8 \\
        DPNET \cite{DPNET} (ACM MM' 2023) & 76.7 $\pm$ 4.3 & 72.7 $\pm$ 3.7 & 73.8 $\pm$ 4.0 & 87.8 $\pm$ 1.0 & 88.4 $\pm$ 1.2 & 85.2 $\pm$ 1.2 \\
        DMIB \cite{DMIB} (WACV' 2024) & 77.6 $\pm$ 3.6 & 71.3 $\pm$ 4.3 & 74.9 $\pm$ 3.2 & 86.0 $\pm$ 0.7 & 86.0 $\pm$ 0.9 & 81.6 $\pm$ 0.9 \\
        PCAG \cite{PCAG} (NN' 2024) & 78.1 $\pm$ 3.1 & 73.0 $\pm$ 3.1 & 75.6 $\pm$ 3.1 & 85.2 $\pm$ 1.7 & 85.5 $\pm$ 1.9 & 81.4 $\pm$ 2.6 \\
        GCFANet \cite{GCFANet} (IF' 2024) & 76.5 $\pm$ 3.9 & 73.5 $\pm$ 2.4 & 77.8 $\pm$ 2.7 & 88.6 $\pm$ 1.5 & 88.9 $\pm$ 1.6 & 85.3 $\pm$ 1.6 \\
        \hline
        CLIMD & \textbf{78.5 $\pm$ 2.3} & \textbf{75.4 $\pm$ 2.5} & \textbf{78.5 $\pm$ 2.3} & \textbf{89.2 $\pm$ 1.4} & \textbf{89.4 $\pm$ 1.2} & \textbf{86.5 $\pm$ 1.8} \\
        \bottomrule
    \end{tabular}}
    \end{center}
\end{table*}

\subsubsection{Comparing with SOTA Methods}
We compare CLIMD with state-of-the-art MD methods \cite{mogonet, dynamicalfusion, MLCLNet, DPNET, DMIB, GCFANet}, and the results are shown in Table \ref{tab1}. CLIMD achieves the best performance on both datasets, demonstrating the effectiveness of the proposed curriculum metric in evaluating the degree of imbalance among training samples.

\begin{table*}[htbp]
\caption{The comparisons with other methods specifically designed for class imbalance on MLLC and BRCA.}
\label{tab2}
    \begin{center}
    \resizebox{12cm}{!}{
    \begin{tabular}{l|ccc|ccc}
        \toprule
        \multirow{2}{*}{\textbf{Method}} & \multicolumn{3}{c|}{\textbf{MLLC}} & \multicolumn{3}{c}{\textbf{BRCA}} \\
        \cline{2-7}
        & ACC & Weighted F1 & Macro F1 & ACC & Weighted F1 & Macro F1 \\
        \hline
        Random Undersampling & 73.2 $\pm$ 5.1 & 69.3 $\pm$ 5.2 & 71.6 $\pm$ 6.3 & 83.6 $\pm$ 1.4 & 84.2 $\pm$ 0.8 & 81.1 $\pm$ 1.5 \\
        Oversampling (SMOTE \cite{smote}) & 70.2 $\pm$ 6.8 & 70.3 $\pm$ 5.0 & 68.8 $\pm$ 7.4 & 79.7 $\pm$ 1.3 & 80.4 $\pm$ 1.0 & 76.6 $\pm$ 1.6 \\
        Oversampling (VAE) & 74.4 $\pm$ 4.7 & 72.3 $\pm$ 4.3 & 71.6 $\pm$ 5.6 & 82.4 $\pm$ 1.6 & 83.4 $\pm$ 2.1 & 79.2 $\pm$ 1.8 \\
        WCE Loss & 75.5 $\pm$ 4.7 & 74.0 $\pm$ 3.1 & 72.5 $\pm$ 4.3 & 84.8 $\pm$ 2.2 & 85.0 $\pm$ 2.2 & 81.7 $\pm$ 2.5\\
        Focal Loss \cite{focal} & 75.8 $\pm$ 2.8 & 74.7 $\pm$ 3.7 & 75.8 $\pm$ 3.8 & 85.1 $\pm$ 1.2 & 85.6 $\pm$ 0.8 & 82.6 $\pm$ 1.1 \\
        \hline
        CLIMD & \textbf{78.5 $\pm$ 2.3} & \textbf{75.4 $\pm$ 2.5} & \textbf{78.5 $\pm$ 2.3} & \textbf{89.2 $\pm$ 1.4} & \textbf{89.4 $\pm$ 1.2} & \textbf{86.5 $\pm$ 1.8}  \\
        \bottomrule
    \end{tabular}}
    \end{center}
\end{table*}

\subsubsection{Evaluation of Addressing Class Imbalance}
To evaluate the performance of CLIMD in addressing the issue of class imbalance, we compare it with other methods specifically designed for class imbalance, and the results are presented in Table \ref{tab2}. The explanation for each method is as follows: 
(1) \textbf{Random Undersampling:} Reducing the number of majority classes by random downsampling;
(2) \textbf{Oversampling (SMOTE):} Using SMOTE \cite{smote} algorithm to oversample the minority classes;
(3) \textbf{Oversampling (VAE):} Generating new samples by VAE to oversample the minority classes; 
(4) \textbf{Weighted Cross-Entropy (WCE) Loss:} Assigning different weights to different class samples using the WCE loss;
(5) \textbf{Focal Loss:} Dynamically adjusting the contribution of each sample to the loss using focal loss \cite{focal}.
The results demonstrate that CLIMD is more efficient in addressing the issue of class imbalance.

\subsubsection{Improvements to Different Models}
Our proposed CL framework is plug-and-play, which implies its potential applicability to various other MD models to further improve their performance. To substantiate this claim, we conducted experiments based on several state-of-the-art models \cite{dynamicalfusion, DMIB}, and the results are shown in Table \ref{tab3}. It can be seen that after applying CLIMD, the performance of various MD models has improved to varying degrees. Moreover, the improvement is more pronounced for simpler models.

\begin{table*}[htbp]
\caption{The improvement of applying CLIMD on BRCA dataset.}
\label{tab3}
    \begin{center}
    \resizebox{7cm}{!}{
    \begin{tabular}{l|ccc}
        \toprule
        \textbf{Method} & \textbf{ACC} & \textbf{Weighted F1} & \textbf{Macro F1} \\
        \midrule
        Baseline & 83.6 $\pm$ 1.6 & 84.4 $\pm$ 1.5 & 81.3 $\pm$ 2.1 \\
        Baseline (\textit{w/} CLIMD) & 89.2 $\pm$ 1.4 & 89.4 $\pm$ 1.2 & 86.5 $\pm$ 1.8 \\
        \midrule
        Multi-D & 87.7 $\pm$ 0.3 & 88.0 $\pm$ 0.5 & 84.5 $\pm$ 0.5 \\
        Multi-D (\textit{w/} CLIMD) & 89.6 $\pm$ 0.6 & 90.2 $\pm$ 1.0 & 85.4 $\pm$ 0.8 \\
        \midrule
        DMIB & 86.0 $\pm$ 0.7 & 86.0 $\pm$ 0.8 & 81.6 $\pm$ 0.9 \\
        DMIB (\textit{w/} CLIMD) & 88.4 $\pm$ 0.6 & 88.9 $\pm$ 1.1 & 83.8 $\pm$ 1.4 \\
        \bottomrule
    \end{tabular}}
    \end{center}
\end{table*}

\begin{table}[htbp]
\caption{The ablation study of MCM and TSS on BRCA.}
\label{tab4}
    \begin{center}
    \resizebox{11cm}{!}{
    \begin{tabular}{c|ccc|c|ccc}
        \toprule
        \multirow{2}{*}{\textbf{Method}} & \multicolumn{3}{c|}{\textbf{Curriculum Measurer}} & \multirow{2}{*}{\textbf{Training Scheduler}} & \multirow{2}{*}{\textbf{ACC}} & \multirow{2}{*}{\textbf{Weighted F1}} & \multirow{2}{*}{\textbf{Macro F1}} \\
        \cline{2-4}
        & Intra-modal & & Inter-modal & & & & \\
        \hline
        Baseline & - & & - & - & 83.6 $\pm$ 1.6 & 84.4 $\pm$ 1.5 & 81.3 $\pm$ 2.1 \\
        (1) & \checkmark & & - & ours & 86.6 $\pm$ 1.1 & 86.9 $\pm$ 0.7 & 83.4 $\pm$ 1.4 \\
        (2) & - & & \checkmark & ours & 87.5 $\pm$ 1.2 & 88.0 $\pm$ 0.9 & 84.3 $\pm$ 1.5 \\
        (3) & \checkmark & & \checkmark & Baby Step & 85.1 $\pm$ 2.4 & 86.1 $\pm$ 2.1 & 80.2 $\pm$ 2.7 \\
        (4) & \checkmark & & \checkmark & One-Pass & 84.9 $\pm$ 1.6 & 85.6 $\pm$ 1.2 & 79.2 $\pm$ 1.9 \\
        (5) & \checkmark & & \checkmark & Linear & 86.6 $\pm$ 2.1 & 87.3 $\pm$ 1.6 & 84.0 $\pm$ 2.4 \\
        (6) & \checkmark & & \checkmark & Root & 88.3 $\pm$ 1.6 & 88.8 $\pm$ 1.1 & 85.4 $\pm$ 1.6 \\
        (7) & \checkmark & & \checkmark & Geom & 87.7 $\pm$ 1.9 & 88.3 $\pm$ 1.5 & 84.8 $\pm$ 2.2 \\
        CLIMD & \checkmark & & \checkmark & ours & \textbf{89.2 $\pm$ 1.4} & \textbf{89.4 $\pm$ 1.2} & \textbf{86.5 $\pm$ 1.8} \\
        \bottomrule
    \end{tabular}}
    \end{center}
\end{table}

\subsubsection{Ablation Study}
A series of ablation studies are conducted to validate the effectiveness of the proposed MCM and TSS, the results are shown in Table \ref{tab4}.
Methods (1) and (2) only use intra-modal or inter-modal measure, respectively. Methods (3)-(7) employ Baby Step \cite{CL1}, One-Pass \cite{CL1}, Linear, Root \cite{root}, and Geometric Progression \cite{geom} as training schedulers, respectively. The ablation experiments reveal that the optimal results are achieved when both the proposed curriculum measures and training scheduler are utilized.

\section{Conclusion}
In this paper, a novel CL framework is proposed to address the issue of class imbalance in multimodal diagnosis. We first leverage intra-modal confidence and inter-modal complementarity  to measure training difficulty of each multimodal sample. Then, we design a novel scheduler based on class distribution, which aims to lead the model to gradually adapt to the imbalanced class distribution. Extensive experimental results on multiple datasets demonstrate the efficiency of the proposed method. In future work, we will evaluate the proposed method across diverse medical scenarios using additional datasets.

\vspace{3pt}
~\\\textbf{Acknowledgments.} This work is supported in part by the Natural Science Foundation of China (62276116), in part by China Postdoctoral Science Foundation (2024M751192). 
\vspace{3pt}
~\\\textbf{Disclosure of Interests.} The authors have no competing interests to declare that are relevant to the content of this paper.

\bibliographystyle{splncs04}
\bibliography{Paper-4207}

\end{document}